\documentclass{article}

\usepackage{microtype}
\usepackage{graphicx}
\usepackage{subcaption}
\usepackage{booktabs} 
\usepackage{colortbl}
\usepackage{xcolor}
\usepackage{booktabs}
\usepackage{array}
\usepackage{hyperref}

\usepackage[accepted]{icml2026}

\usepackage{amsmath}
\usepackage{amssymb}
\usepackage{mathtools}
\usepackage{amsthm}

\usepackage[capitalize,noabbrev]{cleveref}

\theoremstyle{plain}

\theoremstyle{definition}

\theoremstyle{remark}

\usepackage[textsize=tiny]{todonotes}

\icmltitlerunning{FIRSTPASS: A Multi-Domain Peer Review Dataset}

\begin{document}

\twocolumn[
  \icmltitle{FIRSTPASS: A Multi-Domain, Multi-Round Peer Review Dataset \\ Grounded in Real Editorial Outcomes}



  \icmlsetsymbol{equal}{*}

  \begin{icmlauthorlist}
    \icmlauthor{Prabhjot Singh}{aut,rdi}
    \icmlauthor{Somnath Luitel}{wku}
    \icmlauthor{Manmeet Singh}{wku}
    \icmlauthor{Josh Durkee}{wku}
  \end{icmlauthorlist}

  \icmlaffiliation{aut}{The University of Texas at Austin, Austin, TX, USA}
  \icmlaffiliation{rdi}{RediMinds Inc., USA}
  \icmlaffiliation{wku}{Disaster Science Operations Center, Western Kentucky University, Bowling Green, KY, USA}

  \icmlcorrespondingauthor{Prabhjot Singh}{prabhjot.singh@utexas.edu}

  \icmlkeywords{Machine Learning, ICML}

  \vskip 0.3in
]



\printAffiliationsAndNotice{}  

\begin{abstract}
Scientific peer review datasets have trained AI systems exclusively on Computer Science and Machine Learning venues, producing models that critique ablation studies yet have never seen a biology reviewer demand contamination controls or a chemist question Nuclear Magnetic Resonance (NMR) spectral assignments. We introduce \textsc{FirstPass}, the first large-scale peer review dataset built on complete multi-round editorial dialogues from a multidisciplinary high-impact journal. Curated from \textit{Nature Communications} mandatory transparent peer review (instituted November 2022), \textsc{FirstPass} comprises 3,668 records spanning five scientific domains (biology, chemistry, neuroscience, physics, and earth science), capturing the full iterative structure of scientific validation: initial referee reports, author point-by-point responses, and updated reviewer assessments. Each record carries an outcome label derived directly from editorial decisions (STANDARD for two-round review; EXTENDED for three or more rounds), providing ground truth absent in all prior corpora. An automated audit confirms 100\% content integrity. Expert reviews average 2,155 words, substantially denser than conference venue reviews. All data, parsing pipelines, and evaluation scripts are released to enable reproducible benchmarking of AI scientific judgment across disciplines.
\end{abstract}

\section{Introduction}
Every peer review dataset used to train AI systems draws exclusively from Computer Science and Machine Learning venues. PeerRead~\citep{kang2018peerread} established this precedent with 14.7K drafts from ACL, NIPS, and ICLR; ReviewMT~\citep{tan2025reviewmt} added multi-turn structure; MARG~\citep{darcy2024marg} introduced multi-agent generation. All remain anchored to a single community's norms. A model trained on this data learns to critique ablation studies. It has never seen a biology reviewer demand contamination controls, a chemist question NMR spectral assignments, a neuroscientist challenge motion artifact correction, or an earth scientist dispute a stratigraphic age model. These are not edge cases: they represent the majority of scientific output.

Two deeper failures compound this domain narrowness. First, existing systems treat peer review as a one-shot event. The author-reviewer dialogue (where claims are stress-tested, new data introduced, and positions updated) is the central mechanism of scientific validation and is absent from all prior training corpora. Second, evaluation is circular: systems are graded on whether generated text \emph{resembles} a review, not on whether the critique aligns with what editors actually demanded be changed. A model optimizing for stylistic mimicry is not exercising scientific judgment; it is producing review-shaped text.

We introduce \textsc{FirstPass}: the first peer review dataset that is multidisciplinary by design, iterative by structure, and outcome-grounded by construction, comprising 3,668 multi-round dialogues from \textit{Nature Communications} with outcome labels derived directly from real editorial decisions.

\section{Dataset Construction}

\textbf{Source.} We build on \textit{Nature Communications} (ISSN 2041-1723) for four reasons: mandatory transparent peer review since November 2022 eliminates opt-in bias; multidisciplinary scope spans biology, chemistry, neuroscience, physics, and earth science; CC BY 4.0 licensing permits derivative use; and reviews averaging 2,155 words, substantially denser than conference venue critiques.

\textbf{Collection.} We query the Springer Nature OpenAccess API for papers published January 2023 to December 2025. Article and peer-review PDFs are parsed via Gemini-3.1-flash-lite-preview~\citep{google2024gemini} with engineered extraction prompts and JSON recovery~\citep{jsonrepair2024}, validated against pdfplumber baselines. A record is retained only if all four paper sections (abstract, introduction, methods, results) exceed 20 words each and at least two complete review rounds are present. Discussion and conclusions are deliberately excluded: reviewers primarily critique methodology and results, and including author interpretations would bias models toward echoing conclusions rather than developing independent critical judgment.

\textbf{Labels.} Outcome labels derive from round count: two rounds equals STANDARD; three or more equals EXTENDED. This operationalization reflects editorial assessment of unresolved concerns without dependence on decision letter text, absent in 97.7\% of records. The label is consistent across five disciplines (62.8\%--71.2\% STANDARD), confirming it captures structural patterns of reviewer-author tension rather than single-domain procedural norms.

\textbf{Integrity and Release.} Automated audit of all 3,668 retained records confirms zero hollow files and 100\% content integrity (defined as non-empty peer-review text matching source PDFs across all structured fields). The remaining 2.3\% of candidate records were excluded for failing the minimum section-length threshold or lacking a second complete review round. Domain distribution: biology (741), chemistry (744), neuroscience (739), physics (727), earth science (717). Paper-level 80/10/10 train/validation/test splits, stratified by domain and label, prevent leakage. All parsing code, extraction prompts, and audit scripts are released to enable independent replication.

\section{Statistics and Tasks}

\begin{figure}[h]
  \centering
  \includegraphics[width=\columnwidth]{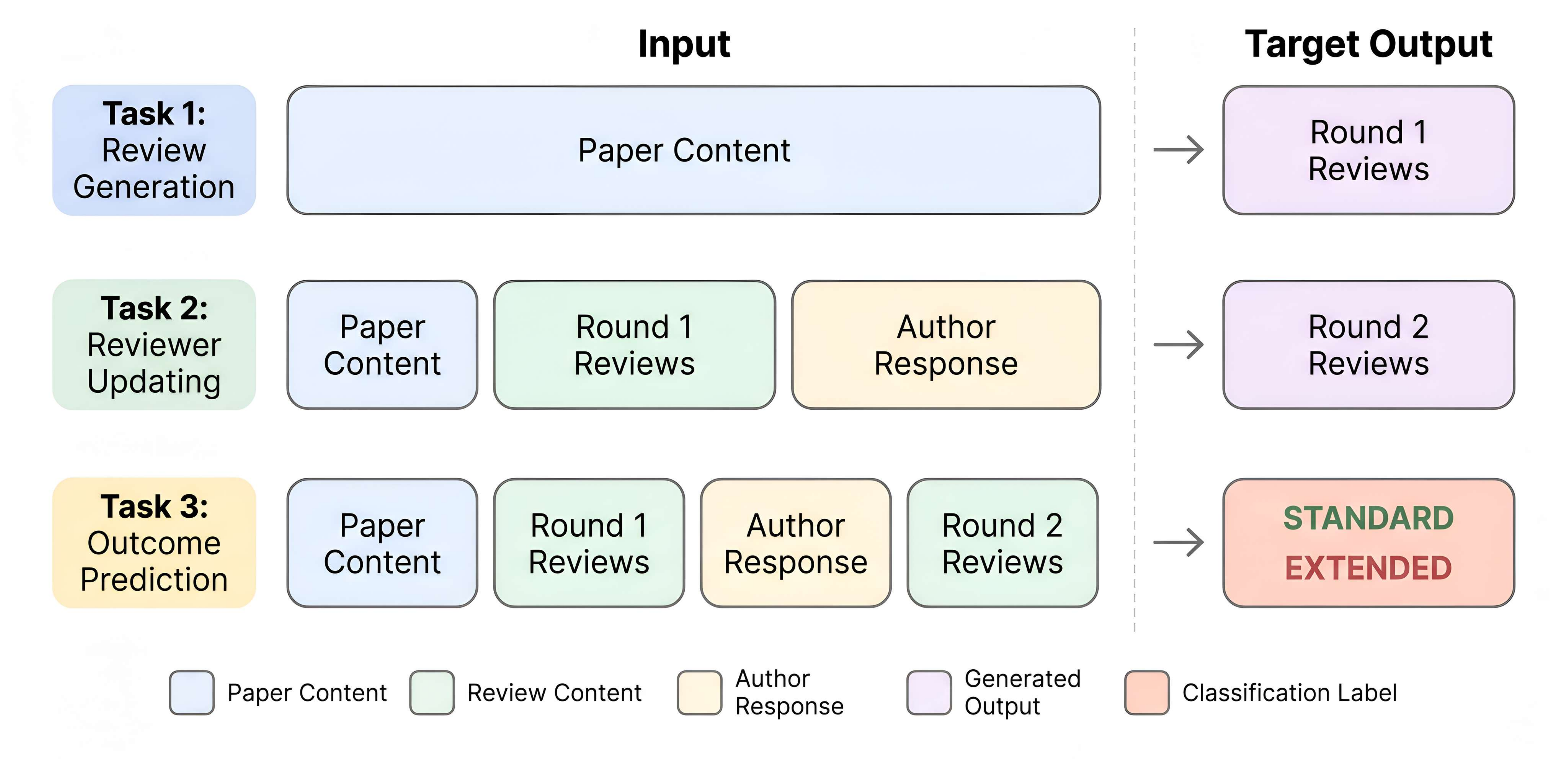}
  \caption{The \textsc{FirstPass} three-task training curriculum.}
  \label{fig:tasks}
\end{figure}

Figure~\ref{fig:tasks} illustrates the three-task curriculum; Table~\ref{tab:stats} reports domain-level statistics. Label balance is consistent across disciplines (62.8\% to 71.2\% STANDARD), confirming stratified splits maintain representative distributions. At 2,155 words average, \textsc{FirstPass} reviews are more than five times longer than ICLR reviews (${\sim}$400 words) and structurally multi-round, providing training signal no prior corpus offers.

\definecolor{firstpassgray}{RGB}{20, 100, 145}
\definecolor{headerblue}{RGB}{219, 234, 254}
\definecolor{totalrow}{RGB}{240, 245, 252}

\begin{table}[h]
\centering
\resizebox{\columnwidth}{!}{%
\begin{tabular}{>{\raggedright\arraybackslash}p{2.2cm} cccccc}
\toprule
\rowcolor{headerblue}
\textbf{Domain} & \textbf{Total} & \textbf{Train} & \textbf{Val} & \textbf{Test} & \textbf{STD (\%)} & \textbf{EXT (\%)} \\
\midrule
Biology      & 741 & 593 & 74 & 74 & 63.2 & 36.8 \\[2pt]
Chemistry    & 744 & 595 & 75 & 74 & 62.8 & 37.2 \\[2pt]
Neuroscience & 739 & 591 & 74 & 74 & 64.6 & 35.4 \\[2pt]
Physics      & 727 & 582 & 73 & 72 & 66.7 & 33.3 \\[2pt]
Earth Sci.   & 717 & 573 & 72 & 72 & 71.2 & 28.8 \\[2pt]
\midrule
\rowcolor{totalrow}
\textcolor{firstpassgray}{\textbf{Total}}
  & \textcolor{firstpassgray}{\textbf{3,668}}
  & \textcolor{firstpassgray}{\textbf{2,934}}
  & \textcolor{firstpassgray}{\textbf{368}}
  & \textcolor{firstpassgray}{\textbf{366}}
  & \textcolor{firstpassgray}{\textbf{65.4}}
  & \textcolor{firstpassgray}{\textbf{34.6}} \\
\bottomrule
\end{tabular}}
\vspace{10pt}
\caption{FIRSTPASS dataset statistics by domain. STD = STANDARD (2 rounds),
EXT = EXTENDED (3+ rounds).}
\vspace{-10pt}
\label{tab:stats}
\end{table}

\textbf{Three-task structure.} Each paper generates three structured training examples targeting distinct scientific judgment capabilities. \textbf{Task 1 (Review Generation):} given paper content, generate Round 1 reviewer comments, training expert critique across five domains. \textbf{Task 2 (Reviewer Updating):} given paper content, Round 1 reviews, and author response, generate Round 2 reviews, training dialogue understanding: which responses are convincing, which introduce new data, which leave methodological debt unresolved. \textbf{Task 3 (Outcome Prediction):} given the full dialogue, predict STANDARD or EXTENDED, the primary evaluation task replacing circular stylistic proxies with real editorial decisions as ground truth.

A fine-tuned Qwen2.5-7B-Instruct~\citep{yang2025qwen} achieves 80.5\% accuracy and F1-macro 78.2\% on Task 3 across all five domains, outperforming Gemini-3.1-flash-lite-preview zero-shot by 10.4 percentage points.

\section{Release and Governance}

\textbf{License and access.} \textsc{FirstPass} inherits CC BY 4.0 from its source material. The dataset, parsing pipeline, extraction prompts, audit scripts, evaluation suite, and fine-tuned model weights are publicly available via Hugging Face Datasets and GitHub\renewcommand{\thefootnote}{\fnsymbol{footnote}}\setcounter{footnote}{1}\footnote{\url{https://github.com/prabhjotschugh/firstpass-peer-review}.}\renewcommand{\thefootnote}{\arabic{footnote}}.

\textbf{Schema and sustainability.} Records are structured JSON; the schema is versioned (v1.0 
at release) with fields covering paper metadata, domain label, four paper sections, separated review rounds, author responses, and outcome label. A unified evaluation script computes accuracy, F1-macro, and F1-EXT, enabling reproducible benchmarking. The pipeline supports annual longitudinal extension as new \textit{Nature Communications} papers publish under mandatory transparent peer review.

\textbf{Privacy.} Reviewer identities are anonymized in the source material (Reviewer 1, 2, etc.) and preserved throughout. No personally identifiable information beyond the published record is included.

\textbf{Governance signal.} \textsc{FirstPass} operationalizes an auditable deployment criterion: a model achieving F1-macro $\geq$ 75\% on Task 3 demonstrates alignment with expert editorial judgment grounded in real outcomes, not stylistic proxies. Beyond peer review, this operationalization applies to any domain where AI judgment must be validated against real expert decisions rather than proxy metrics, including clinical triage, grant evaluation, and regulatory review, making \textsc{FirstPass} a methodological template for outcome-grounded AI governance across scientific institutions. \textsc{FirstPass} provides both the benchmark and the methodology to set that bar. Responsible-use risks include bias replication from a single journal source and potential misuse for synthetic review generation; we release an explicit datasheet documenting known limitations, accepted-papers-only scope, and recommended mitigations.

\newpage

\bibliography{example_paper}
\bibliographystyle{icml2026}

\end{document}